%% file: main.tex
\documentclass[10pt,twocolumn,letterpaper]{article}

\usepackage{cvpr}              %

\input{preamble}

\definecolor{cvprblue}{rgb}{0.21,0.49,0.74}
\usepackage[pagebackref,breaklinks,colorlinks,allcolors=cvprblue]{hyperref}

\def\paperID{****} %
\def\confName{CVPR}
\def\confYear{2001}
\title{KV-Edit: Training-Free Image Editing for Precise Background Preservation}
\author{Tianrui Zhu$^{1}$\footnotemark[1]~, Shiyi Zhang$^{1}$\footnotemark[1]~, Jiawei Shao$^{2}$, Yansong Tang$^{1}$\footnotemark[2]~\\
$^{1}$Shenzhen International Graduate School, Tsinghua University \\
$^{2}$Institute of Artificial Intelligence (TeleAI), China Telecom \\
{\tt\small xilluill070513@gmail.com,sy-zhang23@mails.tsinghua.edu.cn}\\
{\tt\small shaojw2@chinatelecom.cn,tang.yansong@sz.tsinghua.edu.cn}\\
{\small \textbf{\url{https://xilluill.github.io/projectpages/KV-Edit/}}}
}
\begin{document}
\twocolumn[{%
\renewcommand\twocolumn[1][]{#1}%
\maketitle
\begin{center}
    \centering
    \captionsetup{type=figure}
    \includegraphics[width=\linewidth]{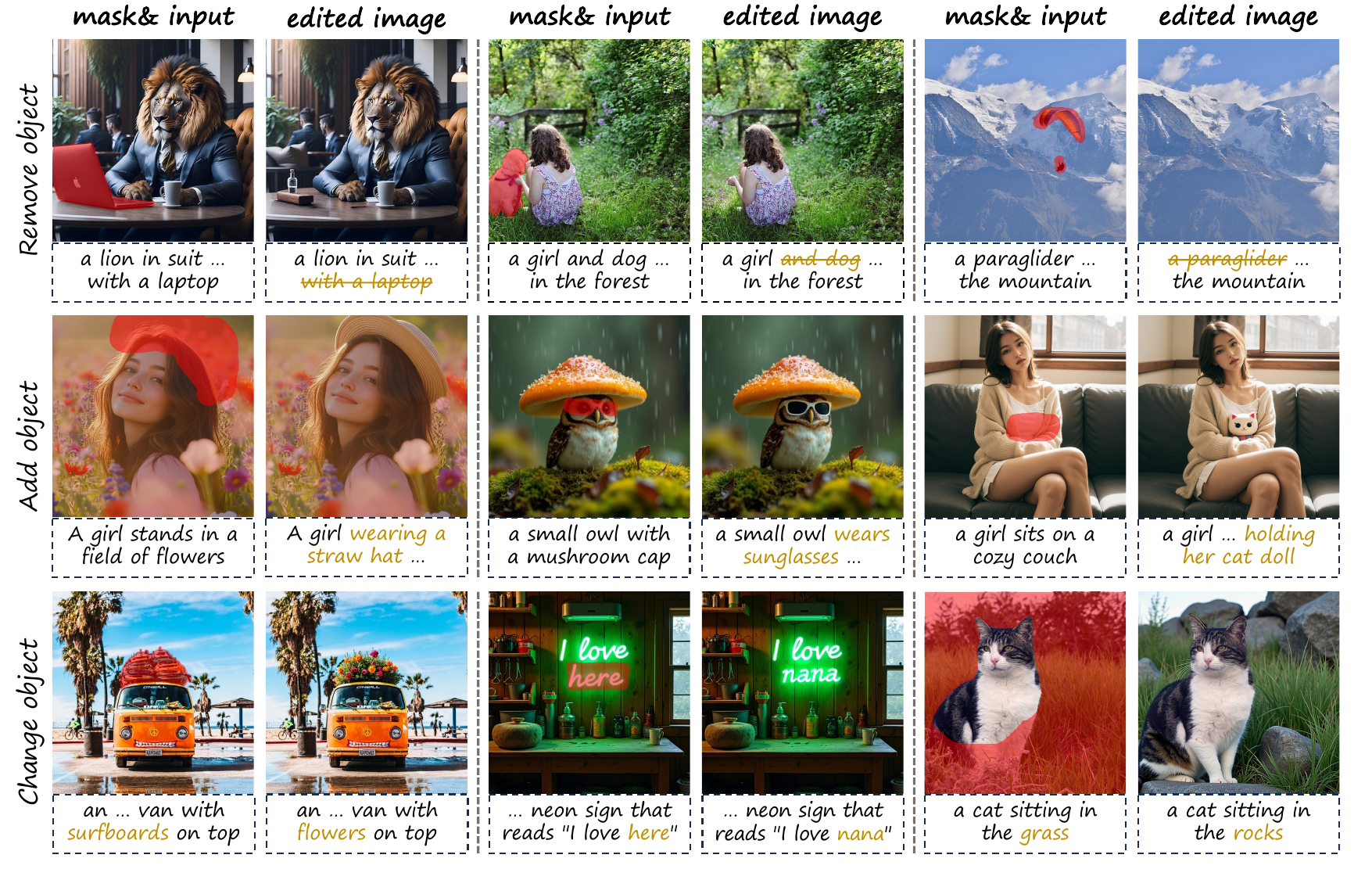}
    \caption{We propose KV-Edit to address the challenge of background preservation in image editing, thereby enhancing the practicality of AI editing. Rather than designing complex mechanisms, we achieve impressive results by simply preserving the key-value pairs of the background. Our method effectively handles common semantic editing operations, including adding, removing, and changing objects.}
    \label{fig:fig1}
\end{center}
}]
\renewcommand{\thefootnote}{\fnsymbol{footnote}}
\footnotetext[1]{Equal contribution.} \footnotetext[2]{Corresponding author.}
\input{sec/0_abstract}    
\input{sec/1_intro}
\input{sec/2_related}
\input{sec/3_method}
\input{sec/4_experiments}

\input{sec/5_conclusion}
{
    \small
    \bibliographystyle{ieeenat_fullname}
    \bibliography{main}
}
\appendix
\renewcommand\thetable{\Alph{table}}
\renewcommand\thefigure{\Alph{figure}}
\renewcommand\thealgorithm{\Alph{algorithm}}

\setcounter{figure}{0}
\setcounter{table}{0}
\setcounter{algorithm}{0}

\clearpage \input{sec/6_appendix}

\end{document}

%% file: preamble.tex
\usepackage{multirow}
\usepackage{algorithm,algpseudocode}
\usepackage{algorithmicx}
\usepackage{stfloats}

%% file: sec/0_abstract.tex
\begin{abstract}
Background consistency remains a significant challenge in image editing tasks. Despite extensive developments, existing works still face a trade-off between maintaining similarity to the original image and generating content that aligns with the target. Here, we propose KV-Edit, a training-free approach that uses KV cache in DiTs to maintain background consistency, where background tokens are preserved rather than regenerated, eliminating the need for complex mechanisms or expensive training, ultimately generating new content that seamlessly integrates with the background within user-provided regions. We further explore the memory consumption of the KV cache during editing and optimize the space complexity to $O(1)$ using an inversion-free method. Our approach is compatible with any DiT-based generative model without additional training. Experiments demonstrate that KV-Edit significantly outperforms existing approaches in terms of both background and image quality, even surpassing training-based methods.
\end{abstract}

%% file: sec/1_intro.tex
\section{Introduction}
\label{sec:intro}

Recent advances in text-to-image (T2I) generation have witnessed a significant shift from UNet~\cite{ronneberger2015u} to DiT~\cite{peebles2023scalable} architectures, and from diffusion models (DMs)~\cite{tang2024post,wang2024towards,dai2024motionlcm} to flow models (FMs)~\cite{flux,kulikov2024flowedit,zhu2024flowie}. Flow-based models, such as Flux~\cite{flux}, construct a straight probability flow from noise to image, enabling faster generation with fewer sampling steps and reduced training resources. DiTs~\cite{peebles2023scalable}, with their pure attention architecture, have demonstrated superior generation quality and enhanced scalability compared to UNet-based models. These T2I models~\cite{rombach2022high,flux,esser2024scaling} can also facilitate image editing, where target images are generated based on source images and modified text prompts.

In the field of image editing, early works~\cite{sdedit,hertz2022prompt,tumanyan2023plug,dong2023prompt} proposed the inversion-denoising paradigm to generate edited images, but they struggle to maintain background consistency during editing. One popular approach is attention modification, such as HeadRouter~\cite{xu2024headrouter} modifying attention maps and PnP~\cite{tumanyan2023plug} injecting original features during the denoising process, aiming to increase similarity with the source image. However, there remains a significant gap between improved similarity and perfect consistency, as it is challenging to control networks' behavior as intended. Another common approach is designing new samplers~\cite{miyake2023negative,mokady2023null} to reduce errors during inversion. Nevertheless, errors can only be reduced but not completely eliminated and both training-free approaches above still require extensive hyperparameter tuning for different cases. Meanwhile, exciting training-based inpainting methods~\cite{li2024brushedit,zhuang2024task} can maintain background consistency but suffer from expensive training costs and potential degradation of quality.

To overcome all the above limitations, we propose a new training-free method that preserves background consistency during editing. Instead of relying on regular attention modification or new inversion samplers for similar results, we implement KV cache in DiTs~\cite{peebles2023scalable} to preserve the key-value pairs of background tokens during inversion and selectively reconstruct only the editing region. Our approach first employs a mask to decouple attention between background and foreground regions and then inverts the image into noise space while caching KV values of background tokens at each timestep and attention layer. During the subsequent denoising process, only foreground tokens are processed, while their keys and values are concatenated with the cached background information. Effectively, we guide the generative model to maintain new content continuity with the background and keep the background content identical to the input. We call this approach \textbf{KV-Edit}.

To further enhance the practical utility of our approach, we conduct an analysis of the removal scenario. This challenge arises from the residual information in surrounding tokens and the object itself which sometimes conflict with the editing instruction. To address this issue, we introduce mask-guided inversion and reinitialization strategies as two enhancement techniques for inversion and denoising separately. These methods further disrupt the information stored in surrounding tokens and self tokens respectively, enabling better alignment with the text prompt. In addition, we apply KV-Edit to the inversion-free method~\cite{xu2024inversion,kulikov2024flowedit}, which no longer caches key-value pairs for all timesteps, but uses KV immediately after one step, significantly reducing the memory consumption of the KV cache.

In summary, our key contributions include:
\textbf{1)} A new training-free editing method that implements KV cache in DiTs, ensuring complete background consistency during editing with minimal hyperparameter tuning.
\textbf{2)} Mask-guided inversion and reinitialization strategies that extend the method's applicability across various editing tasks, offering flexible choices for different user needs.
\textbf{3)} Using the inversion-free method to optimize the memory overhead of our method and enhance its usefulness on PC.
\textbf{4)} Experimental validation demonstrating perfect background preservation while maintaining generation quality comparable to direct T2I synthesis.

%% file: sec/2_related.tex
\input{figs/pipeline}
\section{Related Work}
\label{sec:related}

\subsection{Text-guidanced Editing}

Image editing approaches can be broadly categorized into training-based and training-free methods. Training-based methods~\cite{kawar2023imagic,brooks2023instructpix2pix,ju2024brushnet,li2024brushedit,flux}, have demonstrated impressive editing capabilities through fine-tuning pre-trained generative models on text-image pairs, achieving controlled modifications. Training-free methods have emerged as a flexible alternative, with pioneering works~\cite{sdedit,hertz2022prompt,tumanyan2023plug,dong2023prompt} establishing the two-stage inversion-denoising paradigm. Attention modification has become a prevalent technique in these methods~\cite{cao2023masactrl,li2023stylediffusion,xu2024headrouter,avrahami2024stable,tewel2024add}, specially Add-it~\cite{tewel2024add} broadcast features from inversion to denoising process to maintain source image similarity during editing. Some other work~\cite{mokady2023null,miyake2023negative,ju2024pnp,lin2024schedule,wang2024taming} focused on a better inversion sampler such as the RF-solver~\cite{wang2024taming} designs a second-order sampler. The methods most similar to ours~\cite{avrahami2023blended,chen2024region,liu2024lipe,tewel2024add} attempt to preserve background elements by blending source and target images at specific timesteps using masks. A common consensus is that accurate masks are crucial for better quality, where user-provided inputs~\cite{ju2024brushnet,li2024brushedit} and segmentation models~\cite{ravi2024sam,huang2024segment,yang2022lavt,yang2024language,Liu_2024_CVPR,liu2024universal,bai2024self} prove to be more effective choices compared to masks derived from attention layers in UNet~\cite{ronneberger2015u}. However, the above methods frequently encounter failure cases and struggle to maintain perfect background consistency during editing, while training-based methods~\cite{kawar2023imagic,brooks2023instructpix2pix,zhuang2024task,ju2024brushnet,li2024brushedit,flux} face the additional challenge of computational overhead.

\subsection{KV cache in Attention Models}

KV cache is a widely-adopted optimization technique in Large Language Models  (LLMs)~\cite{brown2020language,xiao2023efficient,bai2023qwen,liu2024deepseek} to improve the efficiency of autoregressive generation. In causal attention, since keys and values remain unchanged during generation, recomputing them leads to redundant resource consumption. KV cache addresses this by storing these intermediate results, allowing the model to reuse key-value pairs from previous tokens during inference. This technique has been successfully implemented in both LLMs~\cite{brown2020language,xiao2023efficient,bai2023qwen,liu2024deepseek} and Vision Language Models (VLMs)~\cite{li2023blip,achiam2023gpt,liu2024visual,zhang2024flash,ye2024voco,huang2024learn,ye2024atp}. However, it has not been explored in image generation and editing tasks, primarily because image tokens are typically assumed to require bidirectional attention~\cite{dosovitskiy2020image,he2022masked}.

%% file: figs/pipeline.tex
\begin{figure*}[htbp]
    \centering
    \includegraphics[width=\linewidth]{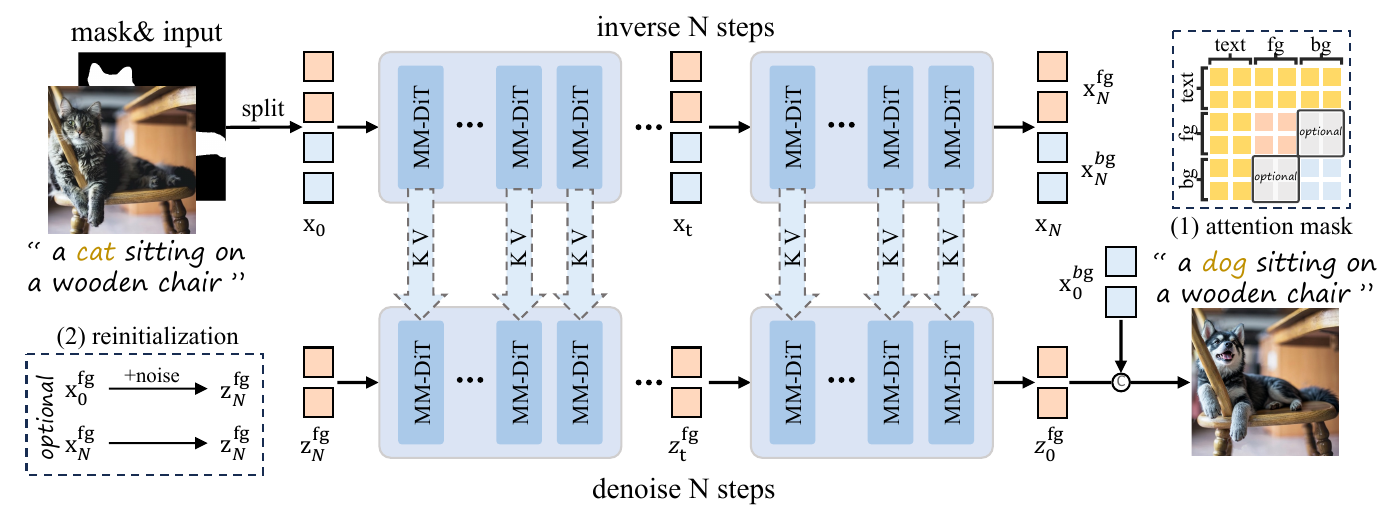}
    \caption{\textbf{Overview of our proposed KV-Edit.} Given an input image and mask, we separate the image into foreground and background. Here, $\mathbf{x}$ and $\mathbf{z}$ denote intermediate results in inversion and denoising processes respectively. Starting from $\mathbf{x}_0$, we first perform inversion to obtain predicted noise $\mathbf{x}_N$ while caching KV pairs. Then, we choose the input $\mathbf{z}^{fg}_N$ and generate edited foreground content $\mathbf{z}^{fg}_0$ based on a new prompt. Finally, we concatenate it with the original background $\mathbf{x}^{bg}_0$ to obtain the edited image with preserved background.}
    \label{fig:pipeline}
    \vspace{-10pt}
\end{figure*}

%% file: sec/3_method.tex
\section{Method}
\label{sec:method}
In this section, we first analyze the reasons why the inversion-denoising paradigm~\cite{dong2023prompt,hertz2022prompt} faces challenges in background preservation. Then, we introduce the proposed KV-Edit method, which achieves strict preservation of background regions during the editing process according to the mask. Finally, we present two optional enhancement techniques and an inversion-free version to improve the usability of our method across diverse scenarios.

\subsection{Preliminaries}

Deterministic diffusion models like DDIM~\cite{ddim} and flow matching~\cite{lipman2022flow} can be modeled using ODE~\cite{song2020score} to describe the probability flow path from noise distribution to real distribution. The model learns to predict velocity vectors that transform Gaussian noise into meaningful images. During the denoising process, $\mathbf{x_1}$ represents noise, $\mathbf{x_0}$ is the final image, and $\mathbf{x_t}$ represents intermediate results.
\begin{equation}
d\mathbf{x}_t=\left(f(\mathbf{x}_t,t)-\frac{1}{2}g^2(t)\nabla_{\mathbf{x}_t}\log p(\mathbf{x}_t)\right)dt,t \in[0,1].
\label{eq:ode}
\end{equation}
where $\mathbf{s}_\theta(\mathbf{x},t)=\nabla_{\mathbf{x}_t}\log p(\mathbf{x}_t)$ predicted by networks. Both DDIM~\cite{ddim} and flow matching~\cite{lipman2022flow} can be viewed as special cases of this ODE function. By setting 
$f(\mathbf{x}_t,t)=\frac{\mathbf{x}_t}{\overline{\alpha}_t}\frac{d\overline{\alpha}_t}{dt}$, 
$g^2(t)=2\overline{\alpha}_t\overline{\beta}_t\frac{d}{dt}\left(\frac{\overline{\beta}_t}{\overline{\alpha}_t}\right)$, and 
$\mathbf{s}_\theta(\mathbf{x},t)=-\frac{\epsilon_\theta(\mathbf{x},t)}{\overline{\beta}_t}$, we obtain the discretized form of DDIM:
\begin{equation}
\mathbf{x}_{t-1}=\bar{\alpha}_{t-1}\left(\frac{\mathbf{x}_t-\bar{\beta}_t\epsilon_\theta(\mathbf{x}_t,t)}{\bar{\alpha}_t}\right)
+\bar{\beta}_{t-1}\epsilon_\theta(\mathbf{x}_t,t)
\end{equation}
Both forward and reverse processes in ODE follow \cref{eq:ode}, describing a reversible path from Gaussian distribution to real distribution. During image editing, this ODE establishes a mapping between noise and real images, where noise can be viewed as an embedding of the image, carrying information about structure, semantics, and appearance.

Recently, Rectified Flow~\cite{liu2022rectified,rectflow} constructs a straight path between noise distribution and real distribution, training a model to fit the velocity field $\mathbf{v}_\theta(\mathbf{x},t)$. This process can be simply described by the ODE:
\begin{equation}
d\mathbf{x}_t=\mathbf{v}_\theta(\mathbf{x},t)dt,t\in[0,1].
\label{eq:4}
\end{equation}
Due to the reversible nature of ODEs, flow-based models can also be used for image editing through inversion and denoising in less timesteps than DDIM~\cite{ddim}.

\input{figs/method_skip}
\subsection{Rethinking the Inversion-Denoising Paradigm}
\label{sec:Rethinking}
The inversion-denoising paradigm views image editing as an inherent capability of generative models without additional training, capable of producing semantically different but visually similar images. However, empirical observations show that this paradigm only achieves similarity rather than perfect consistency in content, leaving a significant gap compared to users' expectations.This section will analyze the reasons for this issue into three factors.

Taking Rectified Flow~\cite{liu2022rectified,rectflow} as an example, based on \cref{eq:4}, we can derive the discretized implementation of inversion and denoising. The model takes the original image $\mathbf{x}_{t_0}$ and Gaussian noise $\mathbf{x}_{t_N}\in\mathcal{N}(0,\boldsymbol{I})$ as path endpoints. Given discrete timesteps $t=\{t_{N},...,t_{0}\}$, the model predictions $\boldsymbol{v}_\theta(C,\mathbf{x}_{t_i},t_i),i\in\{N,\cdots,1\}$, where $\mathbf{x}_{t_i}$ and $\mathbf{z}_{t_i}$ denote intermediate states in inversion and denoising respectively, as described by the following equations:
\begin{equation}
\mathbf{x}_{t_{i}}=\mathbf{x}_{t_{i-1}}+(t_i-t_{i-1})\boldsymbol{v}_\theta(C,\mathbf{x}_{t_i},t_i)
\end{equation}
\begin{equation}
\mathbf{z}_{t_{i-1}}=\mathbf{z}_{t_i}+(t_{i-1}-t_i)\boldsymbol{v}_\theta(C,\mathbf{z}_{t_i},t_i) 
\end{equation}
Ideally, $\mathbf{z}_{t_0}$ should be identity with $\mathbf{x}_{t_0}$ when directly reconstructed from $\mathbf{x}_{t_N}$. However, due to discretization and causality in the inversion process, we can only estimate using $\boldsymbol{v}_\theta(C,\mathbf{X}_{t_{t-1}},t_{t-1}) \approx \boldsymbol{v}_\theta(C,\mathbf{X}_{t_i},t_i)$, introducing cumulative errors. \cref{fig:skip} shows that with a fixed number of timesteps $N$, error accumulation increases as inversion timesteps approach $t_{N}$, preventing accurate reconstruction.

In addition, consistency is affected by condition. We can divide the image into regions we wish to edit $\mathbf{z}_{t_0}^{fg}$ and regions we want to preserve $\mathbf{z}_{t_0}^{bg}$, where ``fg" and ``bg" represent foreground and background respectively. Based on these definitions, the background denoising process is:
\begin{equation}
\boldsymbol{v}_\theta(C,\mathbf{z}_{t_i},t_i)=\boldsymbol{v}_\theta(C,\mathbf{z}_{t_i}^{fg},\mathbf{z}_{t_i}^{bg},t_i)
\end{equation}
\begin{equation}
\mathbf{z}_{t_{i-1}}^{bg}=\mathbf{z}_{t_i}^{bg}+(t_{i-1}-t_i)\boldsymbol{v}_\theta(C,\mathbf{z}_{t_i}^{fg},\mathbf{z}_{t_i}^{bg},t_i) 
\end{equation}
According to these formulas, when generating edited results, the background will be influenced by both the new condition $C$ and new foreground $\mathbf{z}_{t_i}^{fg}$. \cref{fig:change} demonstrates that background regions change when only modifying the prompt or foreground noise. In summary, uncontrollable background changes can be attributed to three factors: error accumulation, new conditions, and new foreground content. In practice, any single element will trigger all three effects simultaneously. Therefore, this paper will present an elegant solution to address all these issues simultaneously.

\input{figs/method_change}
\subsection{Attention Decoupling}
Traditional inversion-denoising paradigms process background and foreground regions simultaneously during denoising, causing undesired background changes in response to foreground and condition modifications. Upon deeper analysis, we observe that in UNet~\cite{ronneberger2015u} architectures, the extensive convolutional networks lead to the fusion of background and foreground information, making it impossible to separate them. However, in DiT~\cite{peebles2023scalable}, which primarily relies on attention blocks~\cite{vaswani2017attention}, allows us to use only foreground tokens as \textbf{queries}, generating foreground content separately and then combined with the background.

Moreover, directly generating foreground tokens often results in discontinuous or incorrect content relative to the background. Therefore, we propose a new attention mechanism where \textbf{queries} contain only foreground information, while \textbf{keys} and \textbf{values} incorporate both foreground and background information. Excluding text tokens, the image-modality self-attention computation can be expressed as:
\begin{equation}
\mathrm{Att}(\mathbf{Q}^{fg},(\mathbf{K}^{fg},\mathbf{K}^{bg}),(\mathbf{V}^{fg},\mathbf{V}^{bg}))=\mathcal{S}(\frac{\mathbf{Q}^{fg}\mathbf{K}^T}{\sqrt{d}})\mathbf{V}
\label{eq:attn}
\end{equation}
where $\mathbf{Q}^{fg}$ represents \textbf{queries} containing only foreground tokens, $(\mathbf{K}^{fg},\mathbf{K}^{bg})$ and $(\mathbf{V}^{fg},\mathbf{V}^{bg})$ denote the concatenation of background and foreground \textbf{keys} and \textbf{values} in their proper order (equivalent to the complete image's \textbf{keys} and \textbf{values}), and $\mathcal{S}$ represents the softmax operation. Notably, compared to conventional attention computations, \cref{eq:attn} only modifies the \textbf{query} component, which is equivalent to performing cropping at both input and output of the attention layer, ensuring seamless integration of the generated content with the background regions.

\input{figs/method_inf}
\subsection{KV-Edit}
\label{section:kv_edit}
Building upon \cref{eq:attn}, achieving background-preserving foreground editing requires providing appropriate key-value pairs for the background. Our core insight is that background tokens' \textbf{keys} and \textbf{values} reflect their deterministic path from image to noise. Therefore, we implement KV cache during the inversion process, as detailed in \cref{algorithm:algorithm1}. This approach records the \textbf{keys} and \textbf{values} at each timestep and block layer along the probability flow path, which are subsequently used during denoising as shown in \cref{algorithm:algorithm2}. We term this complete pipeline ``KV-Edit" as shown in \cref{fig:pipeline} where ``KV" means KV cache.

Unlike other attention injection methods~\cite{tumanyan2023plug,avrahami2024stable,tewel2024add}, KV-Edit only reuses KV for background tokens while regenerating foreground tokens, without requiring specification of particular attention layers or timesteps. Rather than using the source image as injected information, we treat the deterministic background as context and the foreground as content to continue generating, analogous to KV cache in LLMs. Since the background tokens are preserved rather than regenerated, KV-Edit ensures perfect background consistency, effectively circumventing the three influencing factors discussed in \cref{sec:Rethinking}.

Previous works~\cite{hertz2022prompt,dong2023prompt,cao2023masactrl} often fail in object removal tasks when using image captions as guidance, as the original object still aligns with the target prompt. Through our in-depth analysis, we reveal that this issue stems from the residual information of the original object, which persists both in its own tokens and propagates to surrounding tokens through attention mechanisms, ultimately leading the model to reconstruct the original content.

To address the challenge in removing objects, we introduce two enhancement techniques. First, after inversion, we replace $\mathbf{z}_{t_N}$ with fused noise $\mathbf{z}^{\prime}_{t_N} = \mathrm{noise} \cdot t_N + \mathbf{z}_{t_N}\cdot(1-t_N)$ to disrupt the original content information. Second, we incorporate an attention mask during the inversion process, as illustrated in \cref{fig:pipeline}, to prevent foreground content from being incorporated into the KV values, further reducing the preservation of original content. These techniques serve as optional enhancements to improve editing capabilities and performances in different scenarios as shown in \cref{fig:fig1}.
\input{algorithm/algorithm1}
\input{algorithm/algorithm2}
\input{figs/experiment_compare}

\subsection{Memory-Efficient Implementation}
Inversion-based methods require storing key-value pairs for N timesteps, which can pose significant memory constraints when working with large-scale generative models (e.g., 12B parameters~\cite{flux}) on personal computers. Fortunately, inspired by~\cite{xu2024inversion,kulikov2024flowedit}, we explore an inversion-free approach. The method performs denoising immediately after each inversion step, computing the vector difference between the two results to derive a probability flow path in the $t_0$ space. This approach allows immediate release of KV cache after use, reducing memory complexity from $O(N)$ to $O(1)$.

However, the inversion-free method may occasionally result in content retention artifacts as shown in \cref{fig:inf} and FlowEdit~\cite{kulikov2024flowedit}. Since our primary focus is investigating background preservation during editing, we leave more discussion about inversion-free in supplementary materials.

%% file: figs/method_skip.tex
\begin{figure}[t]
    \centering
    \includegraphics[width=\linewidth]{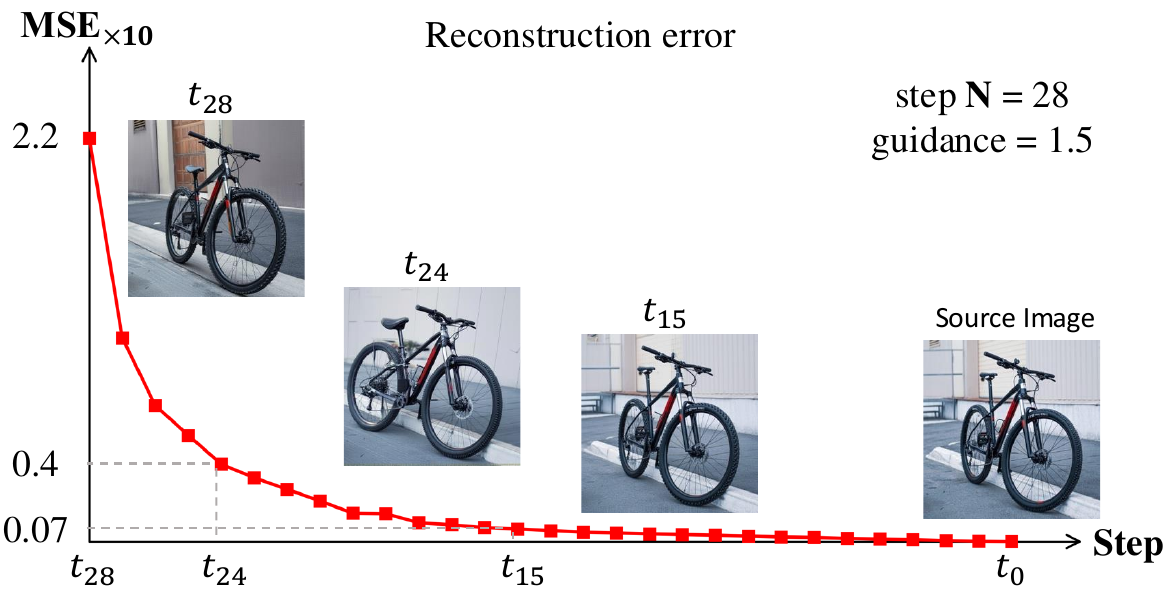}
    \caption{\textbf{The reconstruction error in the inversion-reconstruction process.} Starting from the original image $\mathbf{x}_{t_0}$, the inversion process proceeds to $\mathbf{x}_{t_N}$. During inversion process, we use intermediate images $\mathbf{x}_{t_i}$ to reconstruct the original image and calculate the MSE between the reconstructed image $\mathbf{x}_{t_0}^{\prime}$ and the original image $\mathbf{x}_{t_0}$.}
    \label{fig:skip}
    \vspace{-10pt}
\end{figure}

%% file: figs/method_change.tex
\begin{figure}[t]
    \centering
    \includegraphics[width=\linewidth]{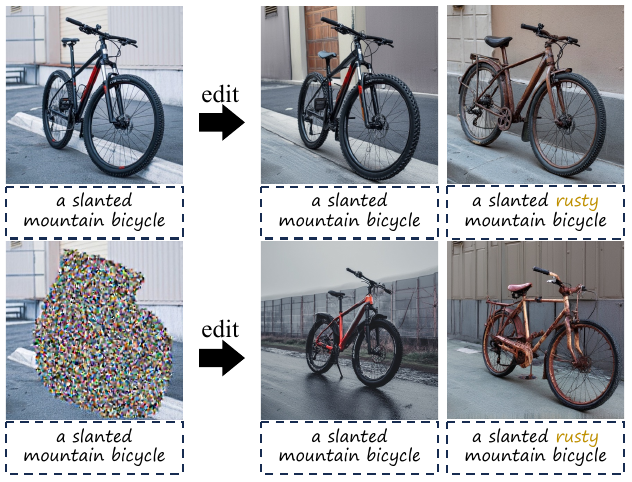}
    \caption{\textbf{Analysis of factors affecting background changes.} The four images on the right demonstrate how foreground content and condition changes influence the final results.}
    \label{fig:change}
    \vspace{-10pt}
\end{figure}

%% file: figs/method_inf.tex
\begin{figure}[t]
    \centering
    \includegraphics[width=\linewidth]{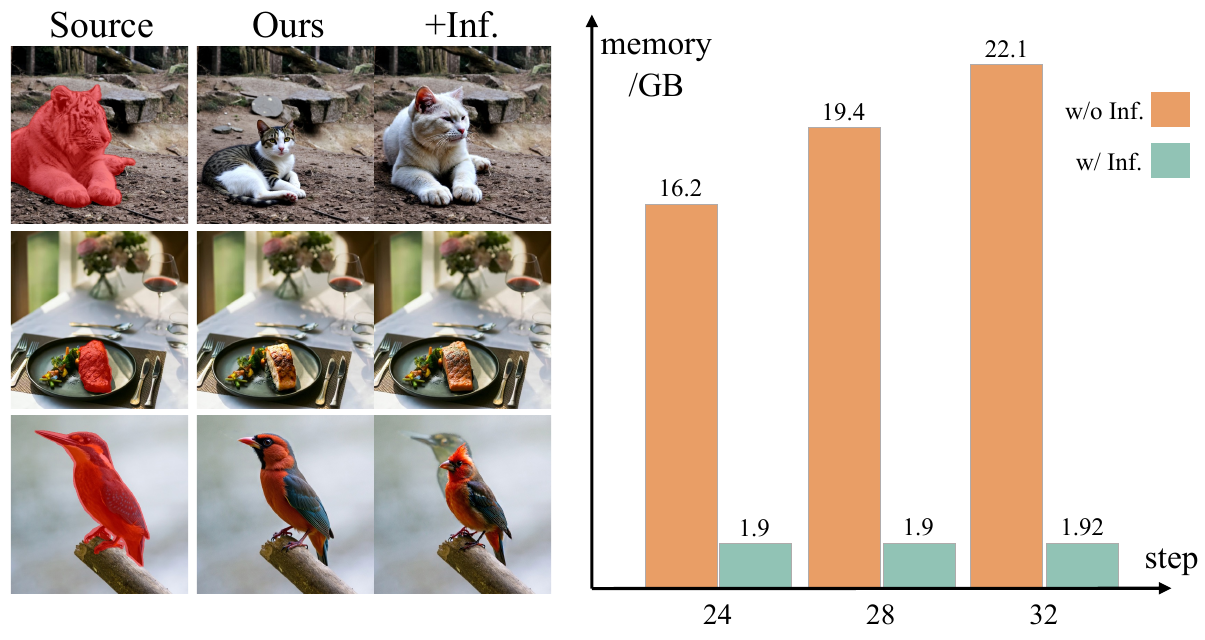}
    \caption{\textbf{Demonstration of inversion-free KV-Edit.} The right panel shows three comparative cases including a failure case, while the left panel illustrates inversion-free approach Significantly optimizes the space complexity to $O(1)$.}
    \label{fig:inf}
    \vspace{-10pt}
\end{figure}

%% file: algorithm/algorithm1.tex
\begin{algorithm}[t]
\caption{Simplified KV cache during inversion}
\begin{algorithmic}[1]
\State \textbf{Input:} $t_{i}$, image $x_{t_{i}}$, $M$-layer block $\{ l_j \}_{j=1}^{M}$, foreground region $mask$, KV cache $C$
\State \textbf{Output:} Prediction vector $V_{\theta t_{i}}$, KV cache $C$
\For{$j = 0$ \textbf{to} $M$}
    \State $Q, K, V = W_Q(x_{t_{i}}), W_K(x_{t_{i}}), W_V(x_{t_{i}})$
    \State $K^{bg}_{ij}, V^{bg}_{ij} = K[1-mask>0], V[1-mask>0]$
    \State $C \gets \text{Append}(K^{bg}_{ij}, V^{bg}_{ij})$
    \State $x_{t_{i}} = x_{t_{i}} + \text{Attn}(Q, K, V)$
\EndFor
\State $V_{\theta t_{i}} = \text{MLP}(x_{t_{i}}, t_{i})$
\State \textbf{Return} $V_{\theta t_{i}}$, $C$
\end{algorithmic}
\label{algorithm:algorithm1}
\end{algorithm}

%% file: algorithm/algorithm2.tex
\begin{algorithm}[t]
\caption{Simplified KV cache during denosing}
\begin{algorithmic}[1]
\State \textbf{Input:} $t_{i}$, foreground $z^{fg}_{t_{i}}$, $M$-layer block $\{ l_j \}_{j=1}^{M}$, KV cache $C$
\State \textbf{Output:} Prediction vector $V^{fg}_{\theta t_{i}}$
\For{$j = 0$ \textbf{to} $M$}
    \State $Q^{fg}, K^{fg}, V^{fg} = W_Q(z^{fg}_{t_{i}}), W_K(z^{fg}_{t_{i}}), W_V(z^{fg}_{t_{i}})$
    \State $K^{bg}_{ij}, V^{bg}_{ij} = C_K[i, j], C_V[i, j]$
    \State $K, V = \text{Concat}(K^{bg}_{ij}, K^{fg}),\text{Concat}(V^{bg}_{ij}, V^{fg})$
    \State $z^{fg}_{t_{i}} = z^{fg}_{t_{i}} + \text{Attn}(Q^{fg}, K, V)$
\EndFor
\State $V^{fg}_{\theta t_{i}} = \text{MLP}(z^{fg}_{t_{i}}, t_{i})$
\State \textbf{Return} $V^{fg}_{\theta t_{i}}$
\end{algorithmic}
\label{algorithm:algorithm2}
\end{algorithm}

%% file: figs/experiment_compare.tex
\begin{figure*}[htbp]
    \centering
    \includegraphics[width=\linewidth]{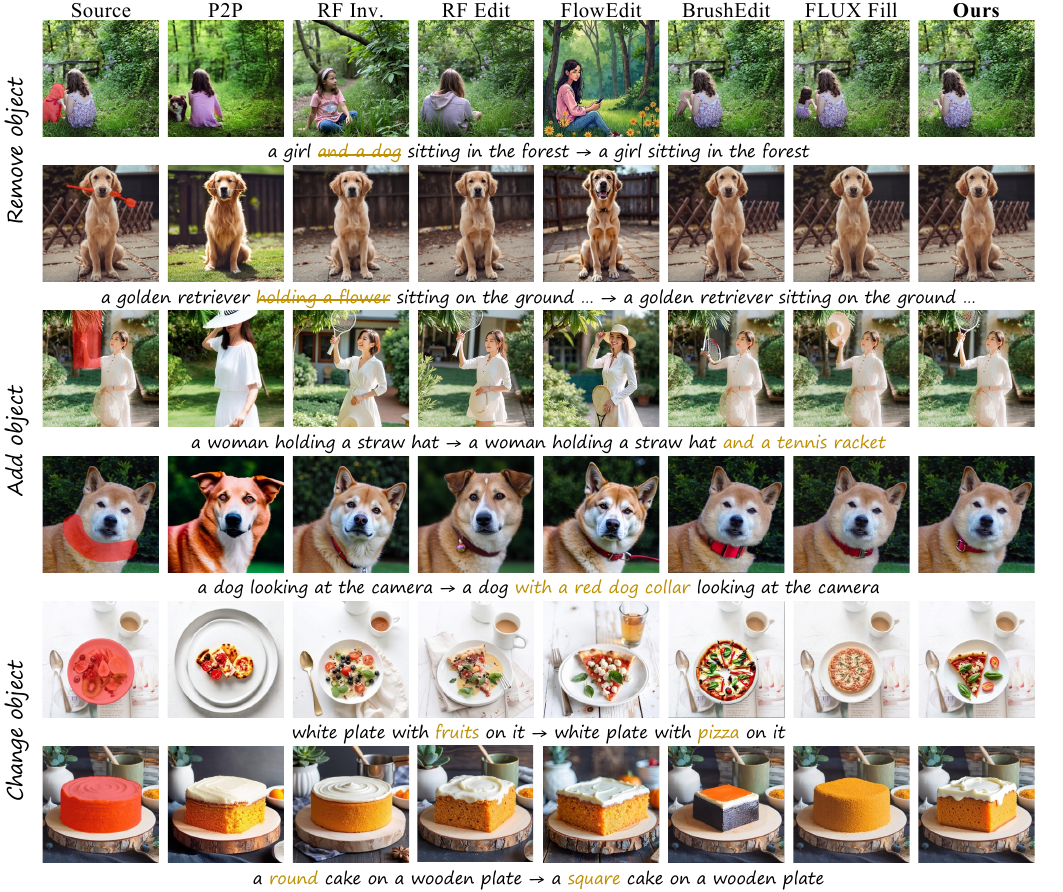}
    \caption{\textbf{Qualitative results on PIE-Bench.} Unlike existing methods, our method demonstrates superior performance by strictly maintaining background consistency and simultaneously following users' text prompt. The comparison also showcases a user-friendly workflow.}
    \label{fig:compare}
    \vspace{-15pt}
\end{figure*}

%% file: sec/4_experiments.tex
\section{Experiments}
\label{sec:experiments}
\input{tables/compare}
\input{tables/ablation_delete}
\subsection{Experimental Setup}

\noindent\textbf{Baselines.} We compare our method against two categories of approaches: (1) Training-free methods including P2P~\cite{hertz2022prompt}, MasaCtrl~\cite{cao2023masactrl} based on DDIM~\cite{ddim}, and RF-Edit~\cite{wang2024taming}, RF-Inversion~\cite{rout2024semantic} based on Rectified Flow~\cite{rectflow}; (2) Training-based methods including BrushEdit~\cite{li2024brushedit} and FLUX-Fill~\cite{flux}, which are based on DDIM and Rectified Flow respectively. In total, we evaluate against six prevalent image editing and inpainting approaches.

\noindent\textbf{Datasets.} We evaluate our method and baselines on nine tasks from PIE-Bench~\cite{ju2024pnp}, which comprises 620 images with corresponding masks and text prompts. Following~\cite{xu2024inversion,li2024brushedit}, we exclude style transfer tasks from PIE-Bench~\cite{ju2024pnp} as our primary focus is background preservation in semantic editing tasks such as object addition, removal, and change.

\noindent\textbf{Implementation Details.} We implement our method based on FLUX.1-[dev]~\cite{flux}, following the same framework as other Rectified Flow-based methods~\cite{wang2024taming,rout2024semantic,kulikov2024flowedit}. We maintain consistent hyperparameters with FlowEdit~\cite{kulikov2024flowedit}, using 28 timesteps in total, skipping the last 4 timesteps ($N=24$) to reduce cumulative errors, and setting guidance values to 1.5 and 5.5 for inversion and denoising processes respectively.
\textbf{NS} in tables and charts represent no skip step ($N=28$). Other baselines retain their default parameters or use previously published results. Unless otherwise specified, ``Ours" in tables refers to the inversion-based KV-Edit without the two optional enhancement techniques proposed in \cref{section:kv_edit}. All experiments are conducted on two NVIDIA 3090 GPUs with 24GB memory.

\noindent\textbf{Metrics} Following~\cite{ju2024brushnet,li2024brushedit,ju2024pnp}, we use seven metrics across three dimensions to evaluate our method. For image quality, we report HPSv2~\cite{zhang2018unreasonable} and aesthetic scores~\cite{schuhmann2022laion}. For background preservation, we measure PSNR~\cite{huynh2008scope}, LPIPS~\cite{zhang2018unreasonable}, and MSE. For text-image alignment, we report CLIP score~\cite{radford2021learning} and Image Reward~\cite{xu2023imagereward}. Notably, while Image Reward was previously used for quality assessment, we found it particularly effective at measuring text-image alignment, providing negative scores for unedited images. Based on this observation, we also utilize Image Reward to evaluate the successful removal of objects.

\subsection{Editing Results}
We conduct experiments on PIE-Bench~\cite{ju2024pnp}, categorizing editing tasks into three major types: removing, adding, and changing objects. For practical applications, these tasks prioritize background preservation and text alignment, followed by overall image quality assessment.

\noindent\textbf{Quantitative Comparison.} \cref{tab:maintable} presents quantitative results including baselines, our method, and our method with the reinitialization strategy. We exclude results with the attention mask strategy, as it shows improvements only in specific cases. Our method surpasses all others in Masked Region Preservation metrics. Notably, as shown in \cref{fig:compare}, methods with PSNR below 30 fail to maintain background consistency, producing results that merely resemble the original. RF-Inversion~\cite{rout2024semantic}, despite obtaining high image quality scores, generates entirely different backgrounds. Our method achieves the third-best image quality, which has been higher than the original images, and perfectly preserving the background at the same time. With the reinitialization process, we achieve optimal text alignment scores, as the injected noise disrupts the original content, enabling more effective editing in certain cases (e.g., object removal and color change). Even compared to training-based inpainting methods~\cite{li2024brushedit,flux}, our approach better preserves backgrounds while following user intentions.

\noindent\textbf{Qualitative Comparison.} \cref{fig:compare} demonstrates our method's performance against previous works across three different tasks. For removal tasks, the examples shown require both enhancement techniques proposed in \cref{section:kv_edit}. Previous training-free methods fail to preserve backgrounds, particularly Flow-Edit~\cite{kulikov2024flowedit} which essentially generates new images despite high quality. Interestingly, training-based methods like BrushEdit~\cite{li2024brushedit} and FLUX-Fill~\cite{flux} exhibit notable phenomena in certain cases (first and third rows in \cref{fig:compare}). BrushEdit~\cite{li2024brushedit}, possibly limited by generative model capabilities, produces meaningless content. FLUX-Fill~\cite{flux} sometimes misinterprets text prompts, generating unreasonable content like duplicate subjects. In contrast, our method demonstrates satisfactory results, successfully generating text-aligned content while preserving backgrounds, eliminating the traditional trade-off between background preservation and foreground editing.

\input{figs/experiment_ablation}
\subsection{Ablation Study}
We conduct ablation studies to illustrate the impact of two enhancement strategies proposed in \cref{section:kv_edit} and the no-skip step on our method's object removal performance. \cref{tab:ablation} presents the results in terms of image quality and text alignment scores. Notably, for text alignment evaluation, we compute the similarity between the generated results and the original prompt using CLIP~\cite{radford2021learning} and Image Reward~\cite{xu2023imagereward} models. This metric proves more discriminative in removal tasks, as still presenting of specific objects in the final images significantly increases the similarity scores.

As shown in \cref{tab:ablation}, the combination of NS (No-skip) and RI (Reinitialization) achieves the optimal text alignment scores. However, we observe a slight decrease in image quality metrics after incorporating these components. We attribute this phenomenon to the presence of too large masks in the benchmark, where no-skip, reinitialization, and attention mask collectively disrupt substantial information, leading to some discontinuities in the generated images. Consequently, these strategies should be viewed as optional enhancements for editing effects rather than universal solutions applicable to all scenarios.

\cref{fig:ablation} visualizes the impact of these strategies. In the majority of cases, reinitialization alone suffices to achieve the desired results, while a small subset of cases requires additional attention masking for enhanced performance.

\subsection{User Study}
We conduct an extensive user study to compare our method with four baselines, including the training-free methods RF-Edit~\cite{wang2024taming}, RF-Inversion~\cite{rout2024semantic}, and the training-based methods BrushEdit~\cite{li2024brushedit} and Flux-Fill~\cite{flux}. We use 110 images from the ``random class" in the PIE-Bench~\cite{ju2024pnp} (excluding style transfer task, images without backgrounds, and controversial content). More than 20 participants are asked to compare each pair of methods based on four criteria: image quality, background preservation, text alignment, and overall satisfaction. As shown in \cref{tab:user_study}, our method significantly outperforms the previous methods, even surpassing Flux-Fill~\cite{flux}, which is the official inpainting model of FLUX~\cite{flux}. Additionally, users' feedback reveals that background preservation plays a crucial role in their final choices, even if RF-Edit~\cite{wang2024taming} achieves high image quality but finally fails in satisfaction comparison. 
\input{tables/user_study}

%% file: tables/compare.tex
\begin{table*}[h]
\begin{center} 
\label{tab:maintable} 
\setlength{\tabcolsep}{3.35mm} %
\begin{tabular}{l|cc|ccc|cc}
\toprule
\multirow{3}{*}[0.8ex]{Method} & \multicolumn{2}{c|}{Image Quality} & \multicolumn{3}{c|}{Masked Region Preservation} &\multicolumn{2}{c}{Text Align} \\
\cmidrule(lr){2-8} & HPS $_{\times 10^2}\uparrow$ & AS $\uparrow$ & PSNR $\uparrow$ & LPIPS $_{\times 10^3}\downarrow$ & MSE $_{\times 10^4}\downarrow$ & CLIP Sim $\uparrow$ & IR$_{\times10}\uparrow$\\
\midrule
VAE$^*$ & 24.93 & 6.37 & 37.65 & 7.93 & 3.86 & 19.69 & -3.65\\
\midrule
P2P~\cite{hertz2022prompt} & 25.40 & 6.27& 17.86 & 208.43 & 219.22 & 22.24 & 0.017 \\
MasaCtrl~\cite{cao2023masactrl} & 23.46 & 5.91 & 22.20 & 105.74 & 86.15 & 20.83 & -1.66\\
RF Inv.~\cite{rout2024semantic} & \underline{27.99} & \textbf{6.74} & 20.20 & 179.73 & 139.85 & 21.71 & 4.34\\
RF Edit~\cite{wang2024taming} & 27.60& \underline{6.56}& 24.44& 113.20& 56.26& 22.08& 5.18\\
BrushEdit~\cite{li2024brushedit} & 25.81 & 6.17 & 32.16 & 17.22 & 8.46 & \underline{22.44} & 3.33\\
FLUX Fill~\cite{flux} & 25.76 & 6.31 & 32.53 & 25.59 & 8.55 & 22.40 & \underline{5.71}\\
\midrule
\textbf{Ours} & 27.21 & 6.49 & \textbf{35.87}& \textbf{9.92}& \textbf{4.69}& 22.39 & 5.63\\
\textbf{+NS+RI} & \textbf{28.05}& 6.40 & \underline{33.30}& \underline{14.80}& \underline{7.45}& \textbf{23.62} & \textbf{9.15}\\
\bottomrule
\end{tabular}
\caption{\textbf{Comparison with previous methods on PIE-Bench.} VAE$^*$ denotes the inherent reconstruction error through direct VAE reconstruction. P2P and MasaCtrl are DDIM-based methods, while RF Inversion and RF Edit are flow-based. BrushEdit and FLUX fill represent training-based methods. \textbf{NS} indicates there is no skip step during inversion. \textbf{RI} indicates the addition of reinitialization strategy. \textbf{Bold} and \underline{underlined} values denote the best and second-best results respectively.}
\end{center}
\vspace{-20pt}
\end{table*}

%% file: tables/ablation_delete.tex
\begin{table}[h]
\begin{center} 
\setlength{\tabcolsep}{0.75mm} %
\begin{tabular}{l|cc|cc}
\toprule
\multirow{3}{*}[0.8ex]{Method} & \multicolumn{2}{c|}{Image Quality} & \multicolumn{2}{c}{Text Align} \\
\cmidrule(lr){2-5} & HPS $_{\times 10^2}\uparrow$ & AS $\uparrow$ & CLIP Sim $^*\downarrow$ & IR$^*_{\times10}\downarrow$\\
\midrule
KV Edit (ours) & 26.76& \textbf{6.49} & 25.50& 6.87\\
+NS & \textbf{26.93}& 6.37& 25.05& 3.17\\
+NS+AM & 26.72& 6.35& 25.00& 2.55\\
+NS+RI & 26.73& 6.34& \textbf{24.82}& \textbf{0.22}\\
+NS+AM+RI & 26.51& 6.28& 24.90& 0.90\\
\bottomrule
\end{tabular}
\caption{\textbf{Ablation study for object removal task.} CLIP Sim$^*$ and IR$^*$ represent alignment between source prompt and new image through CLIP~\cite{radford2021learning} and Image Reward~\cite{xu2023imagereward} to evaluate whether remove particular object from image. \textbf{NS} indicates there is no skip step during inversion. \textbf{RI} indicates the addition of reinitialization strategy. \textbf{AM} indicates that using attention mask during inversion.} 
\label{tab:ablation} 
\end{center}
\vspace{-20pt}
\end{table}

%% file: figs/experiment_ablation.tex
\begin{figure}[t]
    \centering
    \includegraphics[width=\linewidth]{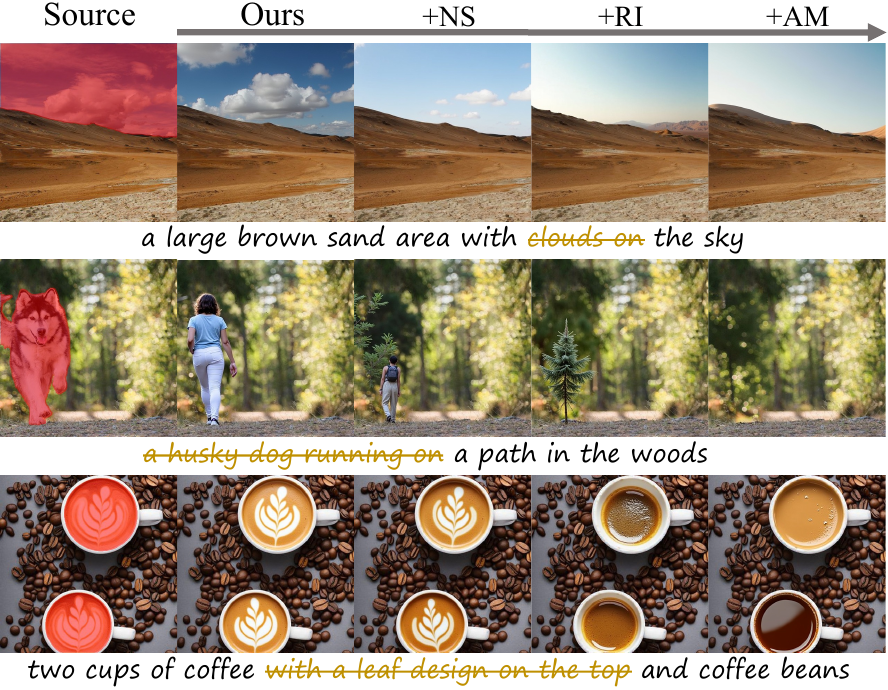}
    \caption{\textbf{Ablation study of different optional strategies on object removal task.} From left to right, applying more strategies leads to stronger removal effect and the right is the best.}
    \label{fig:ablation}
    \vspace{-10pt}
\end{figure}

%% file: tables/user_study.tex
\begin{table}[t]
    \centering
    \small
    \setlength{\tabcolsep}{1.73mm}
        \begin{tabular}{l|ccc|c}
            \toprule

            \multicolumn{1}{l|}{\textbf{ours vs.}} & Quality$\uparrow$ & Background$\uparrow$ &\multicolumn{1}{c|}{Text$\uparrow$}& Overall$\uparrow$\\
            \midrule
            Random$^{*}$ &50.0\% & 50.0\% &50.0\% &50.0\%\\
            \midrule
            RF Inv.~\cite{rout2024semantic}&61.8\% &94.8\% &79.6\% &85.1\%\\
            RF Edit~\cite{wang2024taming}&54.5\% &90.5\% &75.0\% &73.6\%\\
            BrushEdit~\cite{li2024brushedit}&71.8\% & 66.7\% &68.7\% &70.2\%\\
            FLUX Fill~\cite{flux}&60.0\% & 53.7\% &58.6\% &61.9\%\\
            \bottomrule
        \end{tabular}
    \caption{\textbf{User Study.} We compared our method with four popular baselines. Participants were asked to choose their preferred option or indicate if both methods were equally good or not good based on four criteria. We report the win rates of our method compared to baseline excluding equally good or not good instances. Random$^{*}$ denotes the win rate of random choices.}
    \label{tab:user_study}
    \vspace{-10pt}
\end{table}

%% file: sec/5_conclusion.tex
\label{sec:conclusion}
\section{Conclusion}
In this paper, we introduce KV-Edit, a new training-free approach that achieves perfect background preservation in image editing by caching and reusing background key-value pairs. Our method effectively decouples foreground editing from background preservation through attention mechanisms in DiT, while optional enhancement strategies and memory-efficient implementation further improve its practical utility. Extensive experiments demonstrate that our approach surpasses both training-free methods and training-based inpainting models in terms of both background preservation and image quality. Moreover, we hope that this straightforward yet effective mechanism could inspire broader applications, such as video editing, multi-concept personalization, and other scenarios.

%% file: sec/6_appendix.tex
\section*{Appendix}
In this supplementary material, we provide more details and findings. In \cref{app:A}, we present additional experimental results and implementation details of our proposed KV-Edit. \cref{app:B} provides further discussion and data regarding our inversion-free methodology. \cref{app:C} details the design and execution of our user study. Moreover, In \cref{app:D}, we discuss potential future directions and current limitations of our work.
\section{Implementation and More Experiments}
\label{app:A}
\input{figs/app_result_1}
\noindent\textbf{Implementation Details}.
Our code is built on Flux~\cite{flux}, with modifications to both double block and single block to incorporate KV cache through additional function parameters. Input masks are first downsampled using bilinear interpolation, then transformed from single-channel to 64-channel representations following the VAE in Flux~\cite{flux}. In the feature space, the smallest pixel unit is 16 dimensions rather than the entire 64-dimensional token. Therefore, in addition to KV cache, we preserve the intermediate image features at each timestep to ensure fine-grained editing capabilities. In our experiment, inversion and denoising can be performed independently, allowing a single image to be inverted just once and then edited multiple times with different conditions, further enhancing the practicality of this workflow.

\noindent\textbf{Experimental Results}.
Due to space constraints in the main paper, we only present results on the PIE-Bench~\cite{ju2024pnp}. Here, we provide additional examples demonstrating the effectiveness of our approach. To further showcase the flexibility of our method, \cref{fig:app_result_1} and \cref{fig:app_result_2} present various editing target applied to the same source image, without explicitly labeling the input masks because each case corresponds to a different mask. \cref{fig:app_ablation} illustrates the impact of steps and reinitialization strategy on the color changing tasks and inpainting tasks.

When changing colors, as the number of skip-steps decreases and reinitialization strategy is applied, the color information in the tokens is progressively disrupted, ultimately achieving successful results. In our experiments, the optimal number of steps to skip depends on image resolution and content, which can be adjusted based on specific needs and feedback. Unlike previous training-free methods, our approach even can be applied to inpainting tasks after employing reinitialization strategy, as demonstrated in the third row of \cref{fig:app_ablation}. The originally removed regions in inpainting tasks can be considered as black objects, thus requiring reinitialization strategy to eliminate pure black information and generate meaningful content. We plan to further extend our method to inpainting tasks in future work, as there are currently very few training-free methods available for this application.

\noindent\textbf{Attention Scale}
When dealing with large masks (e.g., background changing tasks), our original method may produce discontinuous images including conflicting content, as illustrated in \cref{fig:app_scale}. Stable-Flow~\cite{avrahami2024stable} demonstrated that during image generation with DiT~\cite{peebles2023scalable}, image tokens primarily attend to their local neighborhood rather than globally across most layers and timesteps.

Consequently, although our approach treats the background as a condition to guide new content generation, large masks can introduce generation bias which ignore existing content and generate another objects. Based on this analysis, we propose a potential solution as shown in \cref{fig:app_scale}. We directly increase the attention weights from masked regions to unmasked regions in the attention map (produced by query-key multiplication), effectively mitigating the bias impact. This attention scale mechanism enhances content coherence by strengthening the influence of preserved background on new content.
\input{figs/app_result_2}
\input{figs/app_scale}
\input{figs/app_abaltion}
\section{More Discussions on Inversion-Free}
\label{app:B}
We implement inversion-free editing on Flux~\cite{flux} based on the code provided by FlowEdit~\cite{kulikov2024flowedit}. As noted in FlowEdit~\cite{kulikov2024flowedit}, adding random noise at each editing step may introduce artifacts, a phenomenon we also demonstrate in the main paper. In this section, we primarily explore the impact of inversion-free methods on memory consumption.

\cref{algorithm:algorithm3} demonstrates the implementation of inversion-free KV-Edit, where ``KV-inversion" and ``KV-denoising" refer to single-step noise prediction with KV cache. KV cache is saved during a one-time inversion process and immediately utilized in the denoising process. The final vector can be directly added to the original image without first inversing it to noise. This strategy ensures that the space complexity of KV cache remains $O(1)$ along the time dimension. Moreover, resolution has a more significant impact on memory consumption as the number of image tokens grows at a rate of $O(n^2)$.

We conducted experiments across various resolutions and time steps, reporting memory usage in \cref{tab:app_inf}. When processing high-resolution images and more timesteps, personal computers struggle to accommodate the memory requirements. Nevertheless, we still recommend the inversion-based KV-Edit approach for several reasons:

\begin{enumerate}
    \item Current inversion-free methods occasionally introduce artifacts.
    \item Inversion-based KV-Edit enables multiple editing attempts after a single inversion, significantly improving usability and workflow efficiency.
    \item Large generative models inherently require substantial GPU memory, which presents another challenge for personal computers. Therefore, we position inversion-based KV-Edit as a server-side technology.
\end{enumerate}
\input{tables/app_inf_compare}
\input{algorithm/algorithm3}
\section{User Study Details}
\label{app:C}

We conduct our user study in a questionnaire format to collect user preferences for different methods. We observe that in most cases, users struggle to distinguish the background effects of training-based inpainting methods (e.g., FLUX-Fill~\cite{flux} sometimes increases grayscale tones in images). Therefore, we allowed participants to select ``equally good" regarding background quality.

Additionally, PIE-Bench~\cite{ju2024pnp} contains several challenging cases where all methods fail to complete the editing tasks satisfactorily. Consequently, we allow users to select ``neither is good" for text alignment and overall satisfaction metrics, as illustrated in \cref{fig:app_user}.

We implement a single-blind mechanism where the corresponding method for each question is randomly sampled, ensuring fairness in the comparison. We collect over 2,000 comparison results and calculate our method's win rate after excluding cases where both methods are rated equally.
\input{figs/app_user.tex}

\section{Limitations and Future Work}
\label{app:D}

In this section, we outline the current challenges faced by our method and potential future improvements. While our approach effectively preserves background content, it struggles to maintain foreground details. As shown in \cref{fig:app_ablation}, when editing garment colors, clothing appearance features may be lost, such as the style, print or pleats. 

Typically, during the generation process, early steps determine the object's outline and color, with specific details and appearance emerging later. In the contrast, during inversion, customized object details are disrupted first and subsequently influenced by new content during denoising. This represents a common challenge in the inversion-denoising paradigm~\cite{hertz2022prompt,tumanyan2023plug,dong2023prompt}.

In future work, we could employ trainable tokens to preserve desired appearance information during inversion and inject it during denoising, still without fine-tuning of the base generative model. Furthermore, our method could be adapted to other modalities, such as video and audio editing, image inpainting tasks. We hope that ``KV cache for editing" can be considered an inherent feature of the DiT~\cite{peebles2023scalable} architecture.

%% file: figs/app_result_1.tex
\begin{figure*}[b]
    \centering
    \includegraphics[width=\linewidth]{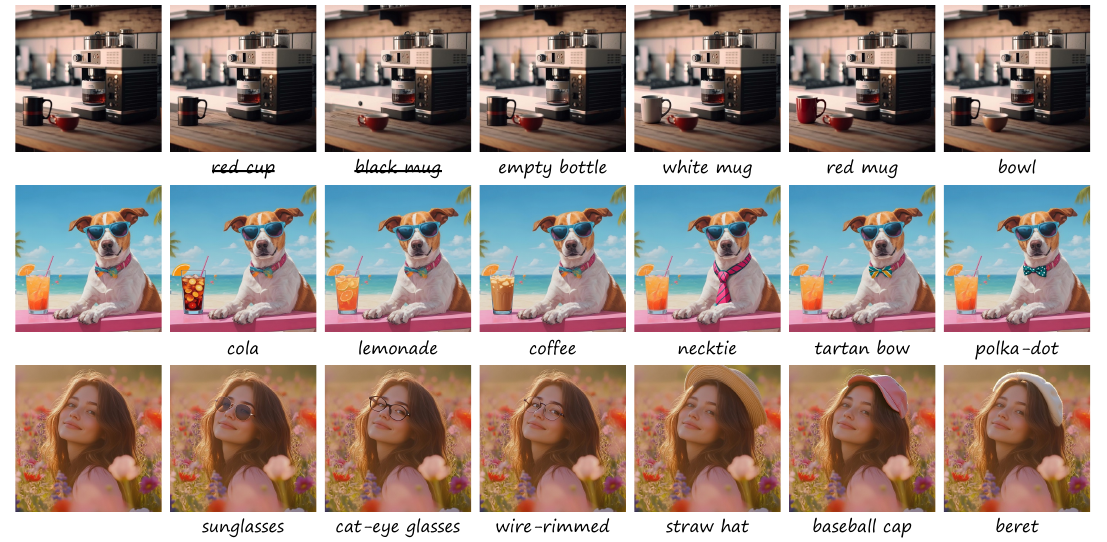}
    \caption{\textbf{Additional editing results of KV-Edit}. Our method demonstrates robust performance with strict background preservation and high image quality. Users can achieve creative designs by simply adjusting text prompts and masks according to their needs.}
    \label{fig:app_result_1}
\end{figure*}

%% file: figs/app_result_2.tex
\begin{figure*}[t]
    \centering
    \includegraphics[width=\linewidth]{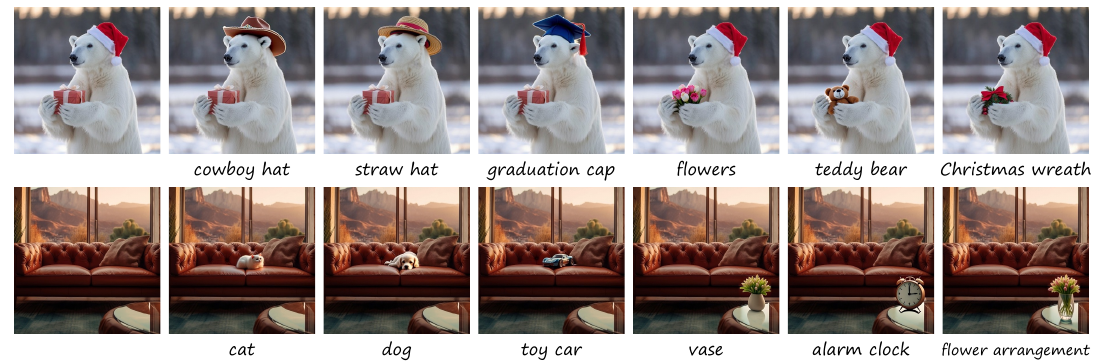}
    \caption{\textbf{Additional editing results of KV-Edit}. Our method demonstrates robust performance with strict background preservation and high image quality. Users can achieve creative designs by simply adjusting text prompts and masks according to their needs.}
    \label{fig:app_result_2}
    \vspace{-15pt}
\end{figure*}

%% file: figs/app_scale.tex
\begin{figure}[b]
    \centering
    \includegraphics[width=0.99\linewidth]{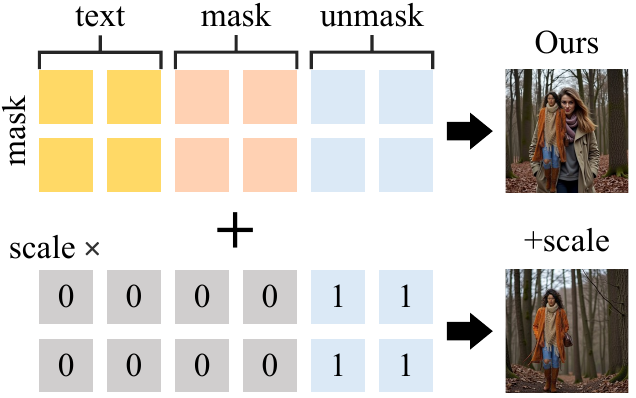}
        \caption{\textbf{Implementation of attention scale}. The scale can be adjusted to achieve optimal results.}
    \label{fig:app_scale}
    \vspace{-10pt}
\end{figure}

%% file: figs/app_abaltion.tex
\begin{figure}[t]
    \centering
    \includegraphics[width=\linewidth]{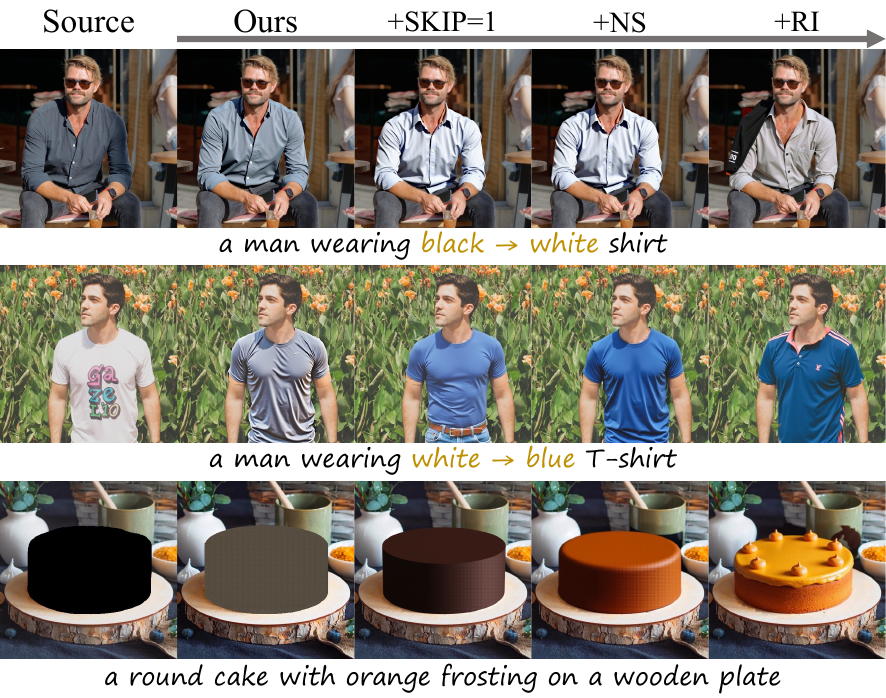}
    \caption{\textbf{Additional ablation studies on two tasks}. The first and second rows demonstrate the impact of timesteps and reinitialization strategy (\textbf{RI}) on color changing. The third row demonstrates the impact of timesteps and \textbf{RI} on the inpainting tasks.}
    \label{fig:app_ablation}
    \vspace{-10pt}
\end{figure}

%% file: tables/app_inf_compare.tex
\begin{table}[t]
    \begin{center} 
    \setlength{\tabcolsep}{4mm} %
    \begin{tabular}{l|cc|cc}
    \toprule
    \multirow{3}{*}[0.8ex]{timesteps} & \multicolumn{2}{c|}{$512 \times 512$} & \multicolumn{2}{c}{$768 \times 768$} \\
    \cmidrule(lr){2-5} & Ours & +Inf. & Ours & +Inf.\\
    \midrule
    24 steps & 16.2G& \textbf{1.9G} & 65.8G& 3.5G\\
    28 steps & 19.4G & \textbf{1.9G}& 75.6G & 3.5G\\
    32 steps & 22.1G& \textbf{1.9G} & 86.5G & 3.5G\\
    \bottomrule
    \end{tabular}
    \caption{\textbf{Memory usage at different resolutions and timesteps}. Our approach has a space complexity of $O(n)$ along the time dimension, while inversion-free methods achieve $O(1)$.}
    \label{tab:app_inf} 
    \end{center}
    \end{table}

%% file: algorithm/algorithm3.tex
\begin{algorithm}[t]
\caption{Simplified Inf. version KV-Edit}
\begin{algorithmic}[1]
\State \textbf{Input:} $t_{i}$, real image $x^{src}_0$, foreground $z^{fg}_{t_{i}}$,foreground region $mask$, KV cache $C$
\State \textbf{Output:} Prediction vector $V^{fg}_{\theta t_{i}}$
\State $N_{t_{i}}\sim\mathcal{N}(0,1)$
\State $x^{src}_{t_i} =(1-t_i)x^{src}_{t_0}+t_i N_{t_{i}}$
\State $V^{src}_{\theta t_{i}},C = \text{KV-Inverison}(x^{src}_{t_i}, t_{i},C)$
\State $\widetilde{z}^{fg}_{t_{i}} = z^{fg}_{t_{i}}+mask\cdot(x^{src}_{t_i}-x^{src}_0)$
\State $\widetilde{V}^{fg}_{\theta t_{i}},C = \text{KV-Denosing}(\widetilde{z}^{fg}_{t_{i}}, t_{i},C)$
\State \textbf{Return} $V^{fg}_{\theta t_{i}} = \widetilde{V}^{fg}_{\theta t_{i}} - V^{src}_{\theta t_{i}}$
\end{algorithmic}
\label{algorithm:algorithm3}
\end{algorithm}

%% file: figs/app_user.tex
\begin{figure}[t]
    \centering
    \includegraphics[width=\linewidth]{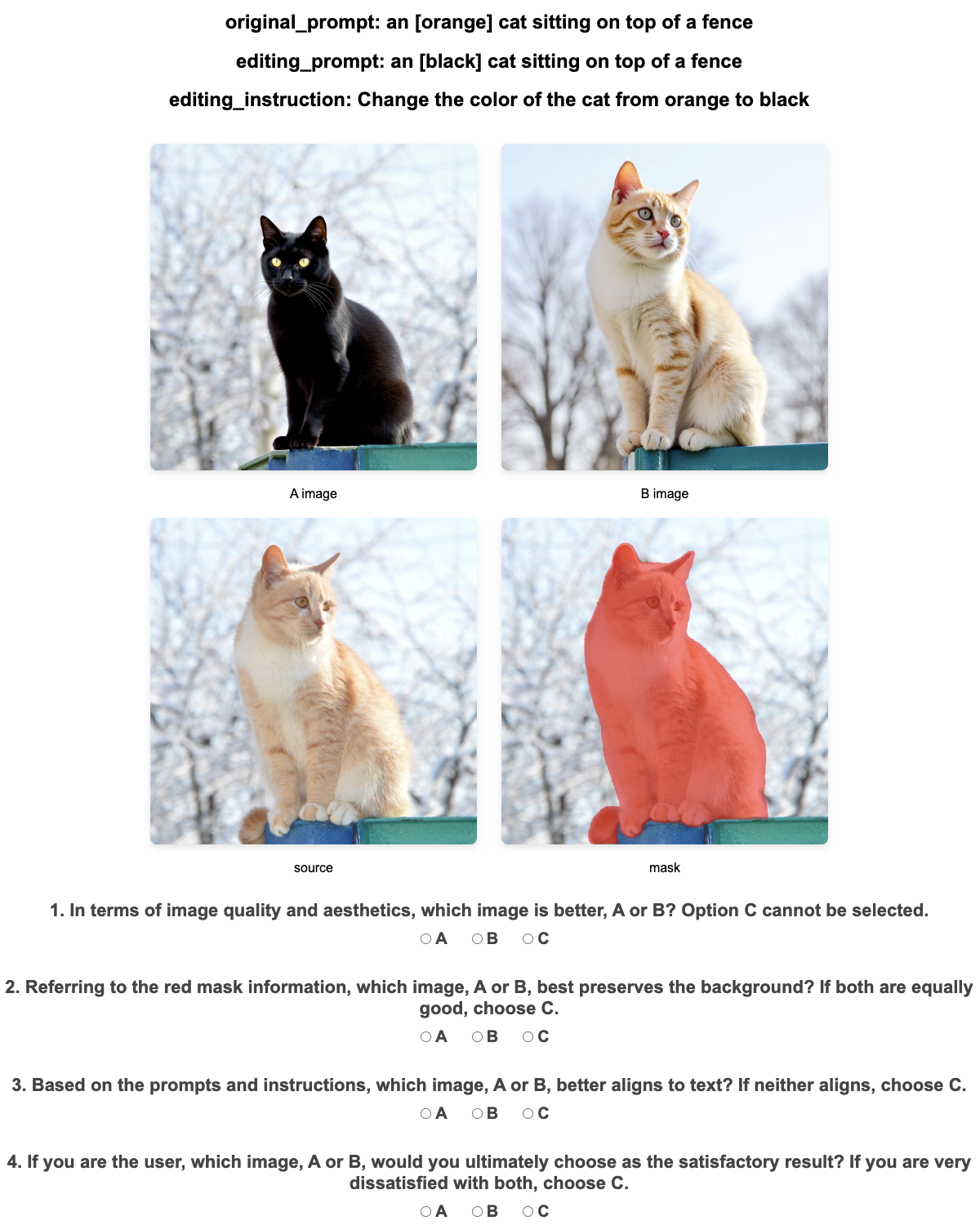}
        \caption{\textbf{User study}. We provide a sample where participants were presented with the original image, editing prompts, results from two different methods for comparison and four questions from four aspects.}
    \label{fig:app_user}
    \vspace{-10pt}
\end{figure}